\RequirePackage[T1]{fontenc}
\documentclass[letterpaper, 10pt, conference]{ieeeconf}
\IEEEoverridecommandlockouts
\usepackage{amsmath,amssymb,amsfonts}
\usepackage{graphicx}
\usepackage{booktabs}
\usepackage{url}
\usepackage[colorlinks=true,linkcolor=blue,citecolor=blue,urlcolor=blue]{hyperref}

\newcommand{\runinhead}[1]{\par\vspace{0.5ex}\textbf{#1:}\ }

\makeatletter
\long\def\@makefntext#1{\parindent 1em\indent\hbox{\@makefnmark}\raggedright #1}
\let\WTRoriginalcaption\@makecaption
\long\def\@makecaption#1#2{%
  \ifx\@captype\@IEEEtablestring
    \WTRoriginalcaption{#1}{#2}%
  \else
    \vskip 2pt\relax
    \parbox[t]{\hsize}{\footnotesize\raggedright\frenchspacing
      \noindent #1.\ #2\par}%
  \fi}
\makeatother

\title{\LARGE \bf
WTR-Drive: World--Task Decomposition with Structured Risk Supervision for End-to-End Driving
}

\author{Haoyu Xing$^{*}$, Shaobing Ding, Guanglei Zhao, Guanhua Zhao, Nawu Peng, and Mingyang Hou%
\thanks{All authors: Voyah Automotive Technology Co., Ltd., Wuhan, China.\newline
$^{*}$Corresponding author: Haoyu Xing (e-mail: 17843105684@163.com).}%
}

\begin{document}

\maketitle
\thispagestyle{empty}
\pagestyle{empty}
\raggedbottom 

\begin{abstract}
Proposal-conditioned scene predictions can lack a common object-level
reference across candidate trajectories. Aggregate planning scores also
leave the objects and times associated with violations implicit.
We present WTR-Drive (World--Task--Risk Driving), a Drive-JEPA-based planner that
separates shared scene estimation from proposal-conditioned reasoning.
A proposal-independent World Memory estimates object states, map structure,
and future motion by re-reading the original visual features at each
refinement round. Proposal feedback updates a residual Task branch without
writing back to World Memory. This branch guides refinement through
candidate-specific retrieval and gated visual feedback. A final Task readout
supplies candidate-conditioned features for scoring.
We derive structured risk targets from existing candidate-scoring rollouts
without additional simulator calls.
For at-fault collision and time-to-collision violations, the targets encode
event existence, association with shared object slots, and the first
violation time bin. On the NAVSIM navtest split, WTR-Drive achieves
94.66 PDMS on v1 and 88.24 EPDMS on v2 (expert-frame sequential evaluation).
These scores are 0.96 and 0.44 points
above the published Drive-JEPA values, respectively. Candidate-level
analysis shows that higher best-candidate scores can coexist with lower
selected-trajectory scores, revealing a gap between candidate quality
and selection effectiveness.
Code is available at
\mbox{\url{https://github.com/xingxing9013605/WTR-Drive}}.
\end{abstract}

\section{INTRODUCTION}

Candidate trajectories represent alternative actions in the same observed
scene. Joint perception--planning systems use intermediate scene
representations~\cite{uniad,vad}. Drive-JEPA combines video pretraining
with multimodal trajectory distillation~\cite{drivejepa}.
iPad~\cite{ipad} retrieves visual features around candidate waypoints.
However, proposal-conditioned predictions
can assign different states to the same object across candidates.
Our approach uses a shared estimate of object states, map structure,
and object motion for all candidates.
Each candidate then reasons about object relevance and potential violations
along its path.

For example, a vehicle may lie outside a lane-keeping trajectory yet
intersect a lane-change trajectory. Aggregate planning scores summarize
trajectory quality but leave the involved object and first violation time
implicit.
Explicit object and time labels supervise candidate--object interactions
and help diagnose object-association and event-timing errors.
iPad and Drive-JEPA already use mapping and collision-related
supervision~\cite{ipad,drivejepa}. We extend this supervision to predict
event existence, association with object slots, and the first violation
time bin for each candidate.

\begin{figure}[!t]
\centering
\includegraphics[width=\columnwidth]{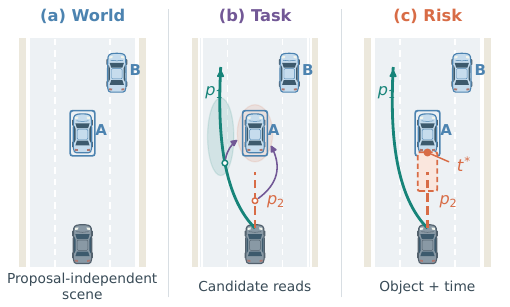}
\caption{World--Task--Risk formulation.
(a) World: all candidates share one scene estimate;
(b) Task: candidates $p_1$ (teal, solid) and $p_2$ (orange, dashed)
read the shared Task Memory.
(c) Risk: only $p_2$ collides with A at $t^{*}$;
the orange footprint marks the ego vehicle at first contact.}
\label{fig:motivation}
\end{figure}

Writing proposal feedback into World Memory would make its updates depend
on the candidate set. We therefore confine this feedback to a residual
Task branch to preserve an observation-based scene estimate for candidate
comparison.
We introduce \textbf{WTR-Drive} (Fig.~\ref{fig:motivation}), a
Drive-JEPA-based planner that separates observation-based World estimation
from proposal-conditioned Task reasoning. Structured risk targets from
existing scoring rollouts supervise candidate--scene interactions.
We make three contributions:

\begin{itemize}
\item \textbf{A proposal-independent World Memory.}
We maintain shared object and map slots that estimate scene structure
and object motion. Each refinement round updates these slots from the
original visual features.

\item \textbf{Residual Task conditioning without World write-back.}
We represent proposal feedback as a residual over the shared World slots.
Task reads and gated visual feedback guide trajectory refinement,
while the final Task readout provides features for scoring.
Proposal feedback never writes back to World Memory.

\item \textbf{Structured risk supervision.}
From existing scoring rollouts, we supervise at-fault collision and
time-to-collision (TTC) event existence, association with World object
slots, and the first violation time bin, without additional simulator calls.
\end{itemize}

On the NAVSIM navtest split~\cite{navsim}, WTR-Drive exceeds the published
Drive-JEPA scores by 0.96 PDMS points on v1 and 0.44 EPDMS points on v2, reaching 94.66 and 88.24.
\label{INTROEND}

\section{RELATED WORK}\label{RELSTART}
\label{sec:rel}

We review scene representations, proposal conditioning, and simulation-based risk supervision.

\subsection{Multimodal Planning and Trajectory Selection}
UniAD~\cite{uniad} and VAD~\cite{vad} connect intermediate scene
representations to planning.
Multimodal planners differ in candidate construction and selection:
Hydra-MDP~\cite{hydramdp} distills human imitation and rule-based scores
over a trajectory vocabulary; DiffusionDrive~\cite{diffusiondrive}
refines multimodal anchors through truncated diffusion;
GoalFlow~\cite{goalflow} combines goal-conditioned flow matching with
goal and trajectory scoring. DriveSuprim~\cite{drivesuprim} couples
progressive filtering with learned scoring.
iPad~\cite{ipad} aligns visual reads with candidate waypoints.
WTR-Drive retains Drive-JEPA's~\cite{drivejepa} pretrained encoder and
multimodal training, while changing the scene features used for
refinement and scoring.

\subsection{Predictive Representations and Driving World Models}
V-JEPA~\cite{vjepa} learns video representations through feature prediction.
Generative world models synthesize future driving video
(GAIA-1~\cite{gaia1}, DriveDreamer~\cite{drivedreamer}, Vista~\cite{vista}),
and OccWorld~\cite{occworld} forecasts 3-D occupancy.
Drive-WM~\cite{drivewm} generates maneuver-conditioned multiview futures
and evaluates plans with image-based rewards.
LAW~\cite{law} predicts future latent features conditioned on ego plans,
while World4Drive~\cite{world4drive} predicts intention-dependent future
latents for candidate evaluation.
WTR-Drive instead estimates shared scene and object motion from
observations, and represents proposal-dependent interactions in Task Memory.

\subsection{Structured Scene Queries and Planning Interaction}
DETR3D~\cite{detr3d} retrieves multi-view features through 3-D references.
BEVFormer~\cite{bevformer} and MapTR~\cite{maptr} organize scene features
through BEV queries and structured map elements, respectively.
SparseDrive~\cite{sparsedrive} links sparse perception, prediction, and
planning, while DriveTransformer~\cite{drivetransformer} allows direct
interaction among agent, map, and planning queries.
WTR-Drive separates World updates from proposal feedback: World slots
re-read the original visual features, and a residual Task branch carries
proposal-dependent interactions without writing back to World Memory.

\subsection{Simulation-Based Supervision and Structured Risk}
nuPlan~\cite{nuplan} evaluates repeated closed-loop
interaction, whereas NAVSIM~\cite{navsim} uses short-horizon non-reactive
simulation. PDM-style criteria~\cite{pdm,navsim} also provide candidate-score
supervision~\cite{hydramdp,drivesuprim}.
DROID~\cite{droid} identifies objects influencing driver behavior through
response supervision and object-removal interventions.
RiskWorld~\cite{riskworld} predicts object risk through latent ego--object
relation rollouts supervised by logged futures.
iPad~\cite{ipad} supervises passability and agent futures at the proposal
level, and Drive-JEPA~\cite{drivejepa} adds mapping and collision targets
to the same proposal-centric pipeline. Both attach these targets to
individual proposals, whereas WTR-Drive assigns them to shared World
slots, so the labels are comparable across candidates.
WTR-Drive associates candidate events with shared World object slots
through one scene-level assignment. Existence, association, and first
violation time bins form auxiliary targets for collision and TTC
violations, reusing scoring rollouts to supervise candidate--scene
relationships.

\section{METHOD}\label{METHSTART}

\begin{figure*}[!t]
\centering
\includegraphics[width=0.90\textwidth]{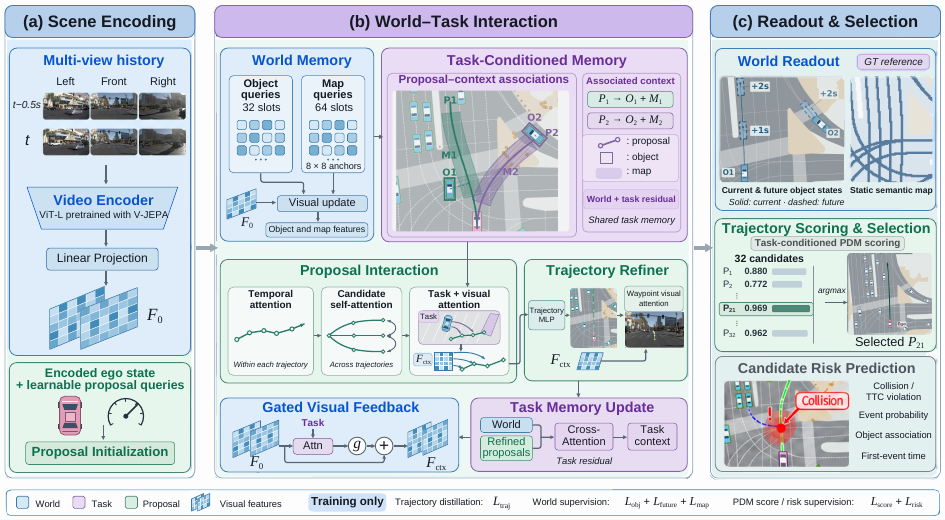}
\caption{Overview of WTR-Drive.
(a) Multi-camera visual encoding and proposal initialization.
(b) In four shared-parameter rounds, World Memory re-reads the original visual features,
while proposal feedback updates Task Memory.
Candidate-specific reads and gated visual feedback carry task information
into refinement without writing back to World Memory,
so it stays proposal-independent.
(c) World readouts estimate scene structure and object motion.
Separate scoring and risk heads predict candidate scores and
event existence, object association, and the first violation time bin.}
\label{fig:overview}
\end{figure*}

\subsection{Problem Setup and Overview}
Given multi-camera images and ego status, we predict $P$ candidate
trajectories with $T$ future poses $(x,y,\theta)$ and select one by its
predicted score. We build on Drive-JEPA~\cite{drivejepa}, which supplies the pretrained
encoder, proposal initialization, trajectory decoder, BEV refiner,
proposal exchange, and multimodal training. Our additions are the
World/Task split over $I$ shared-parameter rounds
(Fig.~\ref{fig:overview}) and the structured risk heads.
Table~\ref{tab:notation} lists the notation;
the batch dimension is omitted.

We write $\operatorname{MHA}_{k}(\mathbf X;\mathbf Y)$ for
multihead-attention output with query input $\mathbf X$
and key/value input $\mathbf Y$.
$\operatorname{Block}_{k}(\mathbf X;\mathbf Y)$ additionally
includes residual connections, layer normalization, and a
feed-forward network.
The subscript identifies the module.
For self-attention, both inputs are the same.

\begin{table}[!t]
\caption{Notation used in the Method.}
\label{tab:notation}
\centering
\small
\setlength{\tabcolsep}{3pt}
\begin{tabular}{@{}ll@{}}
\toprule
Symbol & Meaning (shape for tensors) \\
\midrule
$P,T,I$ & Candidates, forecast steps, rounds \\
$i,p,t,o,r$ & Round, candidate, step, object, risk indices \\
$N,N_v,d$ & Scene slots, visual tokens, channels \\
$\mathbf F_0,\mathbf F_{\mathrm{ctx}}^{(i)}$
& Original / Task-conditioned vision, $N_v\times d$ \\
$\mathbf Q^{(i)},\widehat{\mathbf Q}$
& Proposal / readout features, $P\times T\times d$ \\
$\mathbf Y^{(i)}$
& Trajectory coordinates, $P\times T\times3$ \\
$\mathbf W^{(i)},\mathbf T^{(i)},\boldsymbol\Delta^{(i)}$
& World / Task / residual, $N\times d$ \\
$\mathbf M^{(i)}$
& Memory read by proposals, $N\times d$ \\
\bottomrule
\end{tabular}
\end{table}

\subsection{Scene-Query World Modeling}\label{sec:world}
The \emph{World Memory} concatenates object and fixed-grid map slots,
$\mathbf W=[\mathbf W^{\mathrm{obj}};\mathbf W^{\mathrm{map}}]$.
Each round updates
$\mathbf W^{(i)}=\mathcal W(\mathbf W^{(i-1)},\mathbf F_0)$
from the original visual features without direct proposal feedback;
gradients still jointly train the complete model.
Scene-level targets supervise object states, map semantics, and a
time-conditioned object-motion head. Thus, \emph{world modeling} denotes
scene estimation and motion prediction: refinement revisits the same
observations without physical-time advance or a learned latent transition.
Shared object slots anchor correspondence across candidates, whose
attention weights and risk labels can differ.

\runinhead{Visual Reads for World Slots}
We adapt the geometric sampling strategy of
Drive-JEPA~\cite{drivejepa} and iPad~\cite{ipad} to World slots.
Object references follow current World box estimates;
map references use fixed ground-plane BEV cells.
Visible samples from $\mathbf F_0$ are averaged and projected
separately for objects and maps.
The slots then undergo global attention to $\mathbf F_0$
and joint object--map self-attention.

\subsection{Task-Conditioned Proposal Refinement}\label{sec:task}
Task Memory uses the same object/map layout as World Memory but
incorporates the current proposal set. Initialize learned World slots and
ego-conditioned proposal features with $\boldsymbol\Delta^{(0)}=\mathbf0$,
$\mathbf F_{\mathrm{ctx}}^{(0)}=\mathbf F_0$, and no initial coordinates
($\mathbf Y^{(0)}=\varnothing$). After each World update, we form the memory
read by proposals and refine the proposals:
\begin{equation}
\begin{aligned}
\mathbf M^{(i)}
 &= \mathbf W^{(i)}+\boldsymbol\Delta^{(i-1)},\\
(\mathbf Q^{(i)},\mathbf Y^{(i)})
 &= \mathcal R\big(\mathbf Q^{(i-1)},\mathbf Y^{(i-1)};
 \mathbf M^{(i)},\mathbf F_{\mathrm{ctx}}^{(i-1)}\big).
\end{aligned}
\label{eq:proposal-refine}
\end{equation}
The refined proposal features then update Task Memory
and its residual:
\begin{equation}
\begin{aligned}
\mathbf T^{(i)}
 &= \operatorname{Block}_{\mathrm{task}}(\mathbf W^{(i)};\mathbf Q^{(i)}),\\
\boldsymbol\Delta^{(i)}
 &= \mathbf T^{(i)}-\mathbf W^{(i)}.
\end{aligned}
\label{eq:task-update}
\end{equation}
Thus, proposal refinement uses the updated World Memory
and the \emph{previous} Task residual;
Task Memory is updated afterward.

\subsubsection{Proposal Interaction}
For $i>1$, the operator $\mathcal R$ conditions each token on its previous
coordinates $\mathbf Y^{(i-1)}$ through coordinate encodings and local
reads of $\mathbf F_{\mathrm{ctx}}^{(i-1)}$, using Drive-JEPA's
footprint sampling~\cite{drivejepa}. The first round skips this step
because no coordinates exist yet. Temporal self-attention within each candidate yields tokens $\mathbf U$;
across candidates, temporal means exchange information and their updates
are broadcast to all forecast steps:
\begin{equation}
\begin{aligned}
\bar{\mathbf U}_p&=T^{-1}\sum_t\mathbf U_{pt},\\
\mathbf V_{pt}&=\mathbf U_{pt}
 +[\operatorname{Block}_{\mathrm{prop}}(\bar{\mathbf U};\bar{\mathbf U})-\bar{\mathbf U}]_p.
\end{aligned}
\label{eq:proposal-interaction}
\end{equation}
The resulting tokens query $\mathbf M^{(i)}$. The trajectory MLP then
decodes $\mathbf Y^{(i)}$, and the BEV refiner reads visual features at
these coordinates to produce $\mathbf Q^{(i)}$.
Next, $\operatorname{Block}_{\mathrm{task}}$ forms $\mathbf T^{(i)}$ with
World slots as queries and all $PT$ refined tokens as keys/values.

\subsubsection{Gated Feedback and Readout}
For $i<I$, a learned gate $\alpha=\tanh(g)$ controls visual feedback:
\begin{equation}
\mathbf F_{\mathrm{ctx}}^{(i)}=\mathbf F_0+\alpha\,\operatorname{MHA}_{\mathrm{vis}}
 (\mathbf F_0;\mathbf T^{(i)}),
\label{eq:visual-feedback}
\end{equation}
Each update starts from $\mathbf F_0$, rather than accumulating contexts.
A second gate $\beta=\tanh(g_{\mathrm r})$ and a scale $s\in[0,1]$
control the final readout for scoring and risk prediction:
\begin{equation}
\widehat{\mathbf Q}=\mathbf Q^{(I)}+s\beta\,\operatorname{MHA}_{\mathrm{read}}
 (\mathbf Q^{(I)};\mathbf T^{(I)}),
\label{eq:task-readout}
\end{equation}
Both gates start at $0.01$. Training uses $s=1$;
evaluation compares $s\in\{0,1\}$. This intervention can change scores
and selection, but preserves generated coordinates, World/Task updates,
and visual feedback.

\subsection{Structured Risk Supervision}
Existing PDM scoring rollouts supply risk targets without additional
simulator calls. For at-fault collisions and TTC violations, we record
existence and, when available, track identity and the first violation bin
(not the TTC value). A Hungarian assignment using class and box costs
matches ground-truth objects to World slots. The same track-to-slot
mapping applies to every candidate, even when event outcomes or times differ.

For risk type $r\in\{\mathrm{collision},\mathrm{TTC}\}$,
candidate $p$, future step $t$, and object slot $o$, let
$\mathbf q_{prt}=\phi_r(\widehat{\mathbf Q}_{pt})$ and
$\mathbf k_o=\phi_k(\mathbf v_o)$.
Here
$\mathbf v_o=\mathbf W_o^{\mathrm{obj},(I)}
+\phi_\Delta(\boldsymbol\Delta_o^{\mathrm{obj},(I)})$
combines the World slot with its projected Task residual.
The logit is
\begin{equation}
\ell_{prto}
=\frac{\mathbf q_{prt}^{\top}\mathbf k_o}{\sqrt d}
+[f_{\mathrm{geom}}(\mathbf g_{pto})]_r+\log\pi_o.
\label{eq:risk-relation}
\end{equation}
The affine maps include biases, with
$\mathbf q_{prt},\mathbf k_o\in\mathbb R^d$.
A shared MLP $f_{\mathrm{geom}}:\mathbb R^8\to\mathbb R^2$ contributes
one value per risk type; $\pi_o$ is the predicted nonempty-slot probability.
Relative to pose $\mathbf Y^{(I)}_{pt}$, $\mathbf g_{pto}$ contains normalized
longitudinal/lateral offsets and distance in the candidate-heading frame,
sine/cosine of relative heading, normalized object length/width, and future
time in seconds.

We aggregate relation logits over time and objects using log-sum-exp:
\begin{equation}
a_{pro}=\log\sum_t e^{\ell_{prto}},\quad
b_{prt}=\log\sum_o e^{\ell_{prto}}.
\label{eq:risk-marginal}
\end{equation}
Softmax over $o$ in $a_{pro}$ and over $t$ in $b_{prt}$
gives the association and time distributions.

A separate MLP predicts event existence from temporally pooled
features,
$\Pr(e_{pr}=1)=
\sigma\big([f_{\mathrm{ex}}(\max_t\widehat{\mathbf Q}_{pt})]_r\big)$,
where $\max_t$ denotes elementwise temporal max pooling
and $\sigma$ is the sigmoid function.
Existence uses focal binary loss on valid labels;
association and time use cross-entropy on positive events with valid labels.
Association also requires a matched slot. Evaluation accounts for label
coverage. These auxiliary heads do not select trajectories
(Sec.~\ref{sec:inference}).

\subsection{Training Objective}
We jointly fine-tune the encoder and downstream modules with trajectory,
score, scene, and risk losses. Best-of-N trajectory supervision uses human
demonstrations (weight $1$) and up to $K=4$ simulator-scored pseudo-teachers
(weight $0.5$).

Each scoring head directly predicts a final score and five component
scores. Simulator targets supervise these with binary cross-entropy (BCE):
the final-score term has weight $1$, and mean component BCE has weight
$0.1$. Components are no-collision, drivable-area compliance, progress,
TTC, and comfort (C for v1; history comfort, HC, for v2).
V1 averages the losses from its two heads.

Where present, inherited predicted-agent classification, regression,
and area terms retain coefficients $1/0.1/2$.
Auxiliary agent classification, box, and BEV-semantic terms retain $1/0.1/1$.
The added object classification, box, and future-pose terms use
$0.05$, $0.01$, and $0.02$. Dense-map and slot-map terms use $0.05$
and $0.0125$. Risk existence, association, and time terms use
$0.10$, $0.05$, and $0.05$, respectively.

\section{EXPERIMENTS}

\subsection{Evaluation Protocol}\label{sec:eval}
{\looseness=-1
We evaluate on 12,146 NAVSIM navtest scenes~\cite{navsim}
with expert trajectories, reporting PDMS on v1 and EPDMS on v2.
Our v2 score uses Drive-JEPA's expert-frame sequential evaluation
without human penalty filter (HPF) correction.
Published results retain their original protocols.
Our v2 evaluation does not measure full two-stage
pseudo-closed-loop performance.}

Component metrics follow NAVSIM~\cite{navsim}: NC (no at-fault
collisions), DAC (drivable area), DDC (driving direction), TLC (traffic
lights), EP (progress), TTC, C (comfort), LK (lane keeping), HC (history
comfort), and EC (extended comfort).

EPDMS$_{\mathrm H}$ applies HPF correction~\cite{navsim_hpf}. On a
0--1 scale, it multiplies NC, DAC, DDC, and TLC by the weighted mean of
EP, TTC, LK, HC, and EC (weights $5,5,2,2,2$), omitting unavailable
terms, and reports 100 times the mean over non-missing expert-frame
scores, separately from published EPDMS. Differences use unrounded
values; scores show two decimals.

Candidate diagnostics during training use a distinct
512-scene validation subset.

\begin{table*}[!t]
\caption{Planning performance on NAVSIM navtest}
\label{tab:comparison}
\centering
\small
\setlength{\tabcolsep}{3pt}
\setlength{\arrayrulewidth}{0.3pt}
\renewcommand{\arraystretch}{1.12}
\textbf{(a) v1}\par\smallskip
\begin{tabular*}{\textwidth}{@{\extracolsep{\fill}}lcc|ccccc|c@{}}
\toprule
\multicolumn{1}{c}{Method} & Backbone & Input & NC$\uparrow$ & DAC$\uparrow$ & EP$\uparrow$ & TTC$\uparrow$ & C$\uparrow$ & PDMS$\uparrow$ \\
\midrule
DiffusionDrive~\cite{diffusiondrive} & ResNet-34 & C+L & 98.20 & 96.20 & 82.20 & 94.70 & \textbf{100.00} & 88.10 \\
GoalFlow~\cite{goalflow} & V2-99 & C+L & 98.40 & 98.30 & 85.00 & 94.60 & \textbf{100.00} & 90.30 \\
DiffusionDriveV2~\cite{diffusiondrivev2} & ResNet-34 & C+L & 98.30 & 97.90 & 87.50 & 94.80 & 99.90 & 91.20 \\
iPad~\cite{ipad} & ResNet-34 & C & 98.60 & 98.30 & 88.00 & 94.90 & \textbf{100.00} & 91.70 \\
DriveSuprim~\cite{drivesuprim} & ViT-L & C & 98.60 & 98.60 & 91.30 & 95.50 & \textbf{100.00} & 93.50 \\
Drive-JEPA~\cite{drivejepa} & ViT-L & C & 99.10 & 98.20 & 91.40 & 96.20 & 99.90 & 93.70 \\
\midrule
WTR-Drive & ViT-L & C & \textbf{99.21} & \textbf{99.06} & \textbf{92.33} & \textbf{96.52} & 99.89 & \textbf{94.66} \\
\bottomrule
\end{tabular*}

\medskip
\textbf{(b) v2}\par\smallskip
\begin{tabular*}{\textwidth}{@{\extracolsep{\fill}}lcc|ccccccccc|c@{}}
\toprule
\multicolumn{1}{c}{Method} & Backbone & Input & NC$\uparrow$ & DAC$\uparrow$ & DDC$\uparrow$ & TLC$\uparrow$ & EP$\uparrow$ & TTC$\uparrow$ & LK$\uparrow$ & HC$\uparrow$ & EC$\uparrow$ & EPDMS$\uparrow$ \\
\midrule
DiffusionDriveV2~\cite{diffusiondrivev2} & ResNet-34 & C+L & 97.70 & 96.60 & 99.20 & 99.80 & 88.90 & 97.20 & 96.00 & 97.80 & \textbf{91.00} & 85.50 \\
DriveSuprim~\cite{drivesuprim} & ViT-L & C & \textbf{98.40} & \textbf{98.60} & \textbf{99.60} & 99.80 & \textbf{90.50} & \textbf{97.80} & 97.00 & \textbf{98.30} & 78.60 & 87.10 \\
Drive-JEPA~\cite{drivejepa} & ViT-L & C & \textbf{98.40} & \textbf{98.60} & 99.10 & 99.80 & 88.40 & \textbf{97.80} & \textbf{97.60} & 97.90 & 84.80 & 87.80 \\
\midrule
WTR-Drive & ViT-L & C & 98.25 & 98.56 & 99.10 & \textbf{99.92} & 89.86 & 97.78 & 97.57 & 98.04 & 84.62 & \textbf{88.24}$^{\dagger}$ \\
\bottomrule
\end{tabular*}
\par\smallskip
\begin{minipage}{\textwidth}
\small\raggedright\frenchspacing
Scores are on a 0--100 scale; $\uparrow$ indicates higher is better.
Bold marks column-wise maxima, including ties after rounding.
Inputs: C (camera), L (LiDAR). Sec.~\ref{sec:eval} defines the metrics.
Baseline scores come from the cited papers and retain their original evaluation protocols.
$\dagger$Our v2 score uses expert-frame sequential evaluation without the HPF correction.
\end{minipage}
\end{table*}

\subsection{Implementation Details}
A V-JEPA ViT-L encodes two frames from each of the left, front, and right
cameras at $256\times512$ resolution; projection to $d=256$ forms
$\mathbf F_0$. We use $P=32$ candidates, $T=8$ poses at $0.5$\,s intervals,
$I=4$ shared-parameter rounds, $32$ object slots, and $8\times8$ map slots.
Spatial ranges are $[-8,56]\times[-32,32]\times[-2,6]$\,m for scene queries
and $[0,56]\times[-32,32]$\,m for the map.
The optional global visual attention in Proposal Interaction is disabled;
World global attention and BEV-refiner visual attention remain active.

We train for 50 epochs with Adam, a learning rate of $10^{-4}$, a 0.1
backbone multiplier, and mixed precision. The global batch size is 576
($9$ nodes, $16$ devices per node, and $4$ samples per device), with no
gradient accumulation. A plateau schedule monitors the validation
score. V1 clips gradients at 1, whereas v2 uses no gradient clipping. Both
single-seed runs use seed 0; validation selects zero-based epochs 41 and 47
for v1 and v2, respectively.

\phantomsection\label{sec:inference}
\runinhead{Inference Protocol}
V1 averages the probabilities of two scoring heads; v2 uses one.
Sequential v2 inference inherits the momentum correction
$s_{\mathrm{sel}}=(14s_{\mathrm{pred}}+2c_{\mathrm{EC}})/16$ when a valid
preceding trajectory is available, where $c_{\mathrm{EC}}$ is extended
comfort. Otherwise, or during training-time validation, selection uses
$s_{\mathrm{pred}}$ directly. We select the highest-scoring candidate.

\subsection{Comparison with Existing Methods}
Table~\ref{tab:comparison} reports total scores and their components:
panel (a) covers v1, and panel (b) includes methods whose papers report v2.
WTR-Drive obtains 94.66 PDMS and 88.24 EPDMS. The corresponding published
Drive-JEPA values are 93.70 and 87.80. The differences are 0.96 and
0.44 points, respectively. Camera coverage, training, and scene modeling vary between
these systems, so this comparison measures their combined behavior.

{\looseness=-1
On v2, WTR-Drive has higher EP but lower NC (98.25 vs.\ 98.40) and
EC (84.62 vs.\ 84.80) than the published Drive-JEPA results.
The total score therefore reflects a tradeoff
among safety, progress, and temporal consistency. It does not establish
that every aspect of planning benefits from the added representation.}

\subsection{Training Variants and Task Readout Interventions}
Table~\ref{tab:readout} compares five training configurations and two
readout interventions. All retain World--Task refinement and structured
risk supervision, so these results do not isolate its contribution.
The scene-query baseline omits the final Task readout, subscore loss,
and map refinement. WTR-Drive adds readout and subscore loss to this baseline.
The ranking variant adds pairwise ranking loss
to the model without subscore loss.

For each benchmark version, paired On/Off rows share a checkpoint;
only the final Task readout scale $s$ changes (1/0).
World/Task updates, visual feedback, and generated candidates are unchanged.
Recomputed scores can change selection. Both v2 columns aggregate the
same saved trajectories.

\begin{table}[!t]
\caption{Ablation: training variants and Task readout interventions on
NAVSIM navtest. PDMS: v1; EPDMS and EPDMS$_{\mathrm H}$: v2.}
\label{tab:readout}
\centering
\small
\setlength{\tabcolsep}{2pt}
\renewcommand{\arraystretch}{1.12}
\begin{tabular}{@{}p{86pt}crrr@{}}
\toprule
Configuration & Readout & PDMS$\uparrow$ & EPDMS$\uparrow$ & EPDMS$_{\mathrm H}\uparrow$ \\
\midrule
\multicolumn{5}{@{}l}{\emph{Training configurations}} \\
\raggedright Scene-query baseline & -- & 94.20 & 88.32 & 91.59 \\
\raggedright \quad + map refinement & -- & 94.43 & 88.45 & 91.84 \\
\raggedright w/o subscore loss & On & 94.35 & 88.39 & 91.88 \\
\raggedright \quad + ranking loss & On & 94.22 & 88.35 & 91.75 \\
\raggedright WTR-Drive & On & \textbf{94.66} & 88.24 & 91.55 \\
\midrule
\multicolumn{5}{@{}l}{\emph{Inference interventions}} \\
\raggedright w/o subscore loss & Off & 94.48 & \textbf{88.46} & \textbf{91.95} \\
\raggedright WTR-Drive & Off & 94.65 & 87.89 & 91.26 \\
\bottomrule
\end{tabular}
\par\smallskip
\begin{minipage}{\columnwidth}
\small\raggedright\frenchspacing
``+'' adds to the preceding configuration.
On/Off sets the Task readout scale $s$ in \eqref{eq:task-readout} to 1/0;
paired rows share one checkpoint and differ only in $s$.
A dash denotes no readout head.
EPDMS/EPDMS$_{\mathrm H}$ use the same trajectories without/with
HPF correction (Sec.~\ref{sec:eval}).
\end{minipage}
\end{table}

\begin{figure*}[!t]
\centering
\includegraphics[width=\textwidth]{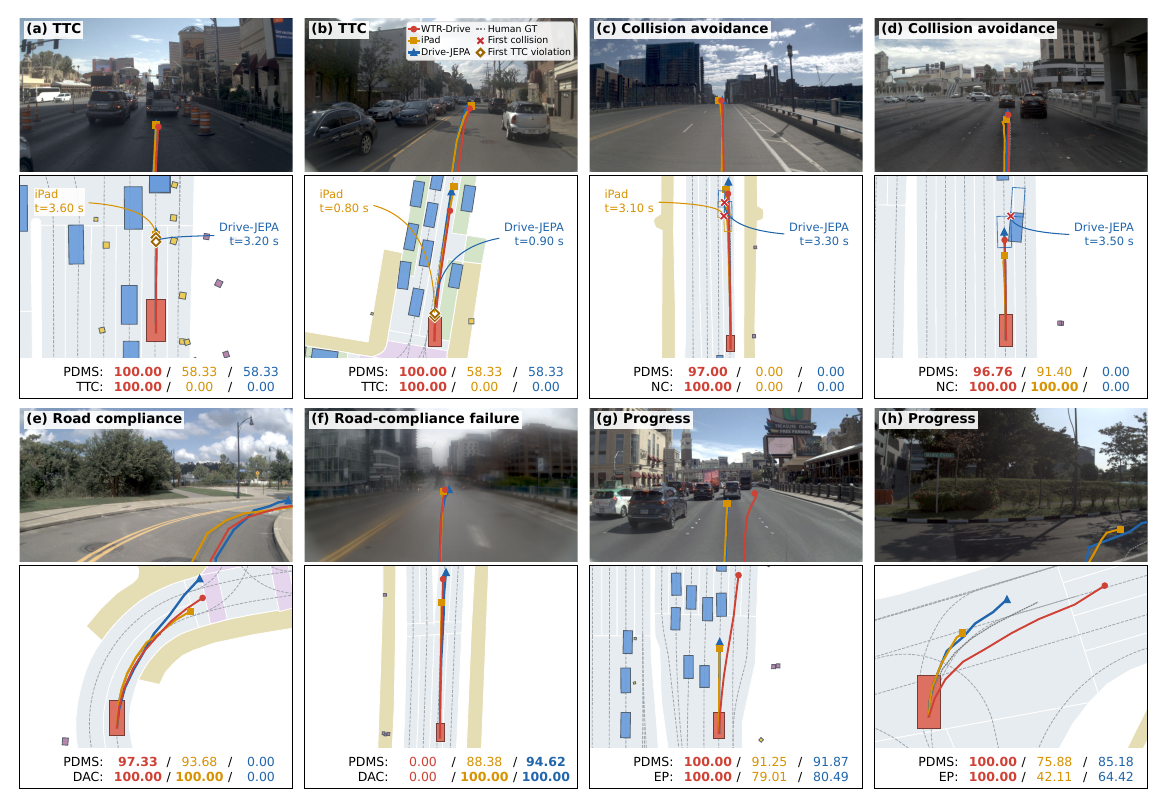}
\caption{Qualitative comparisons on NAVSIM v1, manually selected by
category from 14 paired scenes: seven favorable cases and one failure (f).
Panels report PDMS and one submetric: (a,b) TTC,
(c,d) NC (collision avoidance), (e,f) DAC (road compliance),
(g,h) EP (progress).}
\label{fig:trajectory-comparison}
\end{figure*}

All results below are from single-seed runs. The readout model without
subscore loss (On) exceeds the scene-query baseline by 0.14 PDMS points.
The full model gains 0.31 points over this variant, whereas adding ranking
loss reduces PDMS by 0.12 points. On v2, the full model trails the
variant without subscore loss by 0.15 EPDMS points and 0.33
EPDMS$_{\mathrm H}$ points.

The map variant adds class weighting and convolutional refinement
to the scene-query baseline. It scores 0.23 PDMS points above this baseline,
with higher centerline IoU but lower road IoU; the two changes
are not evaluated separately.

The effect of final Task readout depends on the training configuration.
For the full model, enabling readout changes PDMS by +0.01 points, EPDMS
by +0.35 points, and EPDMS$_{\mathrm H}$ by +0.29 points.
Without subscore supervision, enabling readout yields lower scores on all
three metrics.
These single-seed results do not establish a consistent benefit. Matched
multi-seed ablations are needed to test whether subscore supervision
changes the readout effect.

A separate fixed-learning-rate scene-query run (batch size 192)
achieves 94.45 PDMS and 91.96 EPDMS$_{\mathrm H}$.
Its batch size and learning-rate schedule differ from those of WTR-Drive
(batch size 576), so this comparison cannot isolate an architectural effect.

On the 512-scene subset, the full model achieves a mean best-candidate
score of 96.63 and a mean selected-candidate score of 89.11.
Without subscore supervision, these scores are 96.33 and 89.64,
respectively. A higher best-candidate score therefore does not guarantee
a higher score for the selected trajectory.

These diagnostics use the training scorer, which omits two-frame EC.
The reported v2 evaluation includes EC when available.
This difference merits investigation but does not establish
the cause of the v2 results.

\begin{figure*}[!t]
\centering
\includegraphics[width=0.90\textwidth]{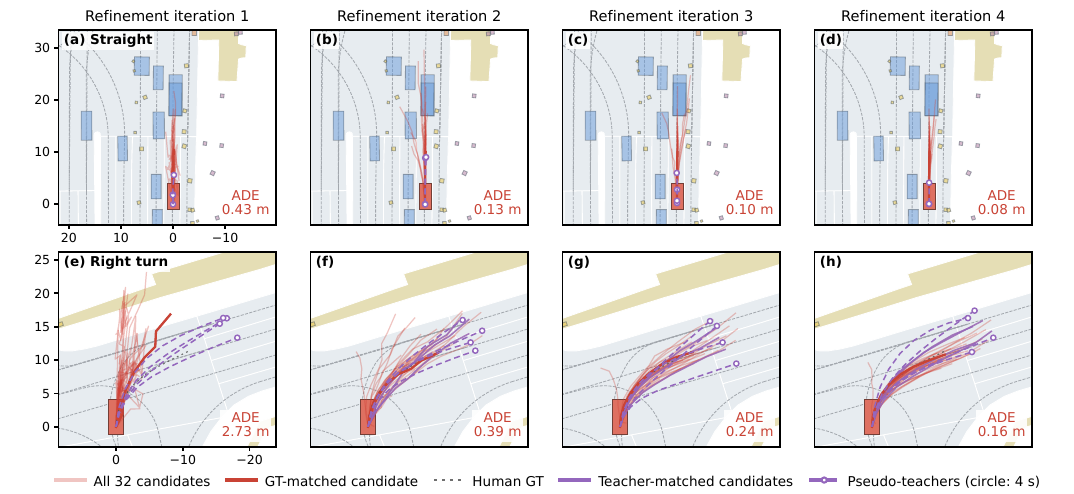}
\caption{Decoder refinement of 32 candidates for two scenes: straight
driving (a)--(d) and a right turn (e)--(h). Red marks the GT-matched trajectory
at each iteration; the annotation reports its average displacement error
(ADE, m) over 4\,s. The matched trajectory may change across iterations
and need not be the score-selected trajectory.}
\label{fig:decoder-refinement}
\end{figure*}

\begin{figure*}[!t]
\centering
\includegraphics[width=\textwidth]{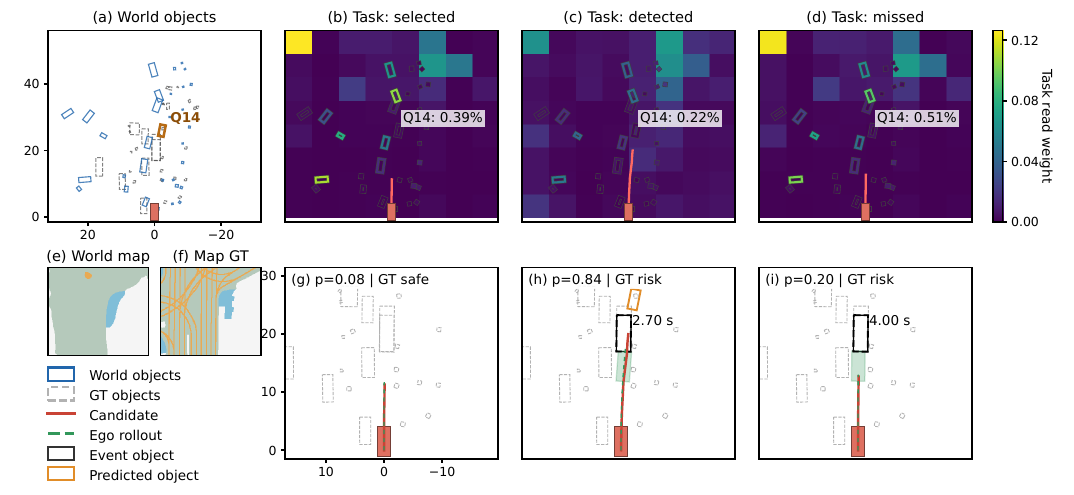}
\caption{World--Task representations and collision diagnostics.
(a) World object estimates.
(b)--(d) Task attention for the selected candidate, a detected
collision, and a missed collision; colors show read weights,
and percentages give the weight on Q14, the same object slot
in (a)--(d).
(e, f) Predicted/ground-truth maps: road (green), walkway (blue),
and centerline (orange).
(g)--(i) Collision probabilities $p$ for the candidates in
(b)--(d), with threshold 0.50; ``GT safe'' and ``GT risk''
denote rollout collision labels.
Solid black boxes mark event objects at the annotated rollout
times; orange boxes mark predicted associations.
BEV axes: lateral/forward distance (m).}
\label{fig:world-task-risk}
\end{figure*}

\subsection{Qualitative Planning and Refinement}
The eight scenes in Fig.~\ref{fig:trajectory-comparison} cover TTC,
collision avoidance, road compliance, and progress. All methods share the
camera frame, calibration, and ego coordinates, and projections assume a
flat ground plane. Each panel reports PDMS and one component score for
WTR-Drive, iPad, and Drive-JEPA, with solid curves for selected
trajectories, gray dashed curves for human GT, filled markers at 4\,s
endpoints, and bold maxima.
WTR-Drive achieves 100.00 PDMS in both TTC cases, compared with 58.33 for
both iPad and Drive-JEPA. In the bridge scene, WTR-Drive achieves 97.00
PDMS, compared with 0.00 for both baselines. In the final turning case,
PDMS is 100.00, 75.88, and 85.18 for WTR-Drive, iPad, and Drive-JEPA,
respectively. Each plotted trajectory is paired with its score from the
same evaluation run.

Replaying the archived trajectories with the native v1 evaluator at
0.10\,s intervals reproduces every reported component score for
panels (a)--(d). In (c), iPad and Drive-JEPA first collide at 3.10 and
3.30\,s; in (d), Drive-JEPA first collides at 3.50\,s. WTR-Drive has no
at-fault collision or TTC violation here. Red crosses mark the overlap
centroid at first contact, amber diamonds the ego rear axle at the first
TTC violation, and annotations give elapsed scene time.

In panel (f), PDMS/DAC are 0.00/0.00 (WTR-Drive), 88.38/100.00 (iPad),
and 94.62/100.00 (Drive-JEPA).
These selected cases illustrate planning differences; they do not estimate
their frequency on the benchmark.

Fig.~\ref{fig:decoder-refinement} follows the full set of 32 candidates
through the four shared decoder iterations. At each iteration, the
highlighted GT match minimizes mean L1 position/heading cost, so its identity
may change. This diagnostic match differs from score-based selection.
Rows use common BEV limits and equal metric scales.

For visualization, we replay the training loss on frozen predictions,
retain the 8,192 scored anchors above 0.95, sample 256 without replacement
when more qualify, and assign up to four pseudo-teachers per iteration
under a fixed seed. This illustrates the procedure but is not a historical
training step.

In the straight scene, Fig.~\ref{fig:decoder-refinement}(a)--(d),
GT-matched ADE decreases from 0.43\,m to 0.08\,m. The score-selected
candidate has ADE values of 2.78\,m and 0.65\,m at iterations 1 and 4.
In the right-turn scene, GT-matched ADE decreases from 2.73\,m to 0.16\,m
across iterations 1--4 (2.73, 0.39, 0.24, and 0.16\,m). The matched
candidate changes from S29 to S14, S26, and S0 under the training
position/heading loss. The final score-selected candidate, S29, has ADE
3.54\,m at iteration 4. Thus, the best GT match improves across rounds,
while score-based selection favors a different candidate. ADE measures XY
similarity to the human trajectory, not simulator safety.

\subsection{World--Task Representations and Structured Risk}
We export World, Task, and risk outputs for all 32 candidates in eight
scenes; selected trajectories and scores match the paired comparison. Risk
diagnostics use the training rollout, scene matching, and validity masks.
Association-label coverage is 71.24\%/74.67\% at the selected v1/v2
epochs, i.e., the fraction of positive candidate--event pairs with known
object identities that map to matched slots, averaged over batches rather
than held-out accuracy.

Fig.~\ref{fig:world-task-risk} shows different responses within one
scene. The predicted map in (e), compared with GT in (f), captures road
regions but misses centerlines. Q14 denotes the same object slot throughout
(zero-based index 14). Under the rollout criteria, the selected trajectory
in (b), (g) has no at-fault collision or TTC violation.
For the collision in (c), (h), the model assigns probability 0.84,
identifies Q14, and predicts the correct time bin. The collision in
(d), (i) has probability 0.20, below threshold 0.50, and an incorrect
object association.
The event footprints come from the label rollout and cannot validate
learned motion forecasts.

Heatmaps and object-edge colors show attention over 64 map and 32 object
slots, normalized jointly and averaged over heads and trajectory times. Q14
receives 0.39\% for the selected plan, 0.22\% for the detected collision,
and 0.51\% for the missed collision, so correct detection occurs at the
smallest weight. Attention indicates feature retrieval, not a calibrated
collision probability.

The visualized failures motivate separate evaluation of localization,
forecasting, association, calibration, and event timing. Matched ablations,
repeated seeds, and efficiency measurements remain outstanding. A matched
risk-loss control would zero the three structured-risk weights while
retaining subscore supervision. Reusing scoring rollouts avoids additional
simulation, though the added modules still incur computation.

\section{CONCLUSION}
WTR-Drive combines shared scene estimation, residual Task conditioning,
and object/time risk supervision within Drive-JEPA, improving on published
Drive-JEPA scores on NAVSIM v1 and v2. Diagnostics distinguish scene
grounding, trajectory refinement, and final selection, and a higher
best-candidate simulator score need not raise the selected score. Future
work will improve centerline geometry, object localization, motion
prediction, and track-to-slot association.

\end{document}